\documentclass{acmtog}

\usepackage{amsfonts}
\usepackage{amssymb}
\usepackage{algorithmic}
\usepackage{algorithm}
\usepackage{amsmath}
\usepackage{graphicx}
\usepackage{caption}
\usepackage{enumitem}
\usepackage{subfigure}
\usepackage{array}
\usepackage{booktabs}	
\usepackage{widetable}
\usepackage{epstopdf}
\usepackage{slashbox}
\usepackage{soul}
\usepackage[table,xcdraw]{xcolor}
\usepackage{mathrsfs}

\usepackage{times}
\usepackage{appendix}

\usepackage{pifont}

\usepackage{multirow}

\usepackage{color}
\definecolor{blue}{rgb}{0,0,1}
\definecolor{red}{rgb}{1,0,0}
\definecolor{orange}{rgb}{0.75, 0.4, 0}

\def\etal{{\textit{et~al.}}}
\DeclareMathOperator*{\argmin}{arg\,min}
\newcommand{\ie}{\textit{i.e.}}
\newcommand{\eg}{\textit{e.g.}}

\acmVolume{VV}
\acmNumber{N}
\acmYear{YYYY}
\acmMonth{Month}
\acmArticleNum{XXX}
\acmdoi{10.1145/XXXXXXX.YYYYYYY}

\begin{document}

\title{Bundle Optimization for Multi-aspect Embedding}

\author{QIONG ZENG {\upshape and} BAOQUAN CHEN
\affil{Shandong University}
\and
YANIR KLEIMAN
\affil{{\'E}cole Polytechnique}
\and
DANIEL COHEN-OR
\affil{Tel Aviv University}
\and
YANGYAN LI
\affil{Shandong University}}



\keywords{Semantic Clustering, Multi-aspect Embedding.}

\acmformat{Qiong Zeng, Baoquan Chen, Yanir Kleiman, Daniel Cohen-Or and Yangyan Li. 2017. Bundle Optimization for Multi-aspect Embedding. ACM Trans. Graph. XX, XX, Article XX, (September 2017), 9 pages.}

\maketitle

\begin{bottomstuff}
Authors' addresses: land and/or email addresses.
\end{bottomstuff}

\maketitle

\begin{abstract}
Understanding semantic similarity among images is the core of a wide range of computer graphics and computer vision applications. An important step towards this goal is to collect and learn human perceptions. Interestingly, the semantic context of images is often ambiguous as images can be perceived with emphasis on different aspects, which may be contradictory to each other.
In this paper, we present a method for learning the semantic similarity among images, inferring their latent aspects and embedding them into multi-spaces corresponding to their semantic aspects.
We consider the multi-embedding problem as an optimization function that evaluates the embedded distances with respect to the qualitative clustering queries. The key idea of our approach is to collect and embed qualitative measures that share the same aspects in bundles. To ensure similarity aspect sharing among multiple measures, image classification queries are presented to, and solved by users. The collected image clusters are then converted into bundles of tuples, which are fed into our bundle optimization algorithm that jointly infers the aspect similarity and multi-aspect embedding. Extensive experimental results show that our approach significantly outperforms state-of-the-art multi-embedding approaches on various datasets, and scales well for large multi-aspect similarity measures.
\end{abstract}

\section {Introduction}
\label{sec:intro}
\begin{figure}[t]
  \centering
    \includegraphics[width=1\linewidth]{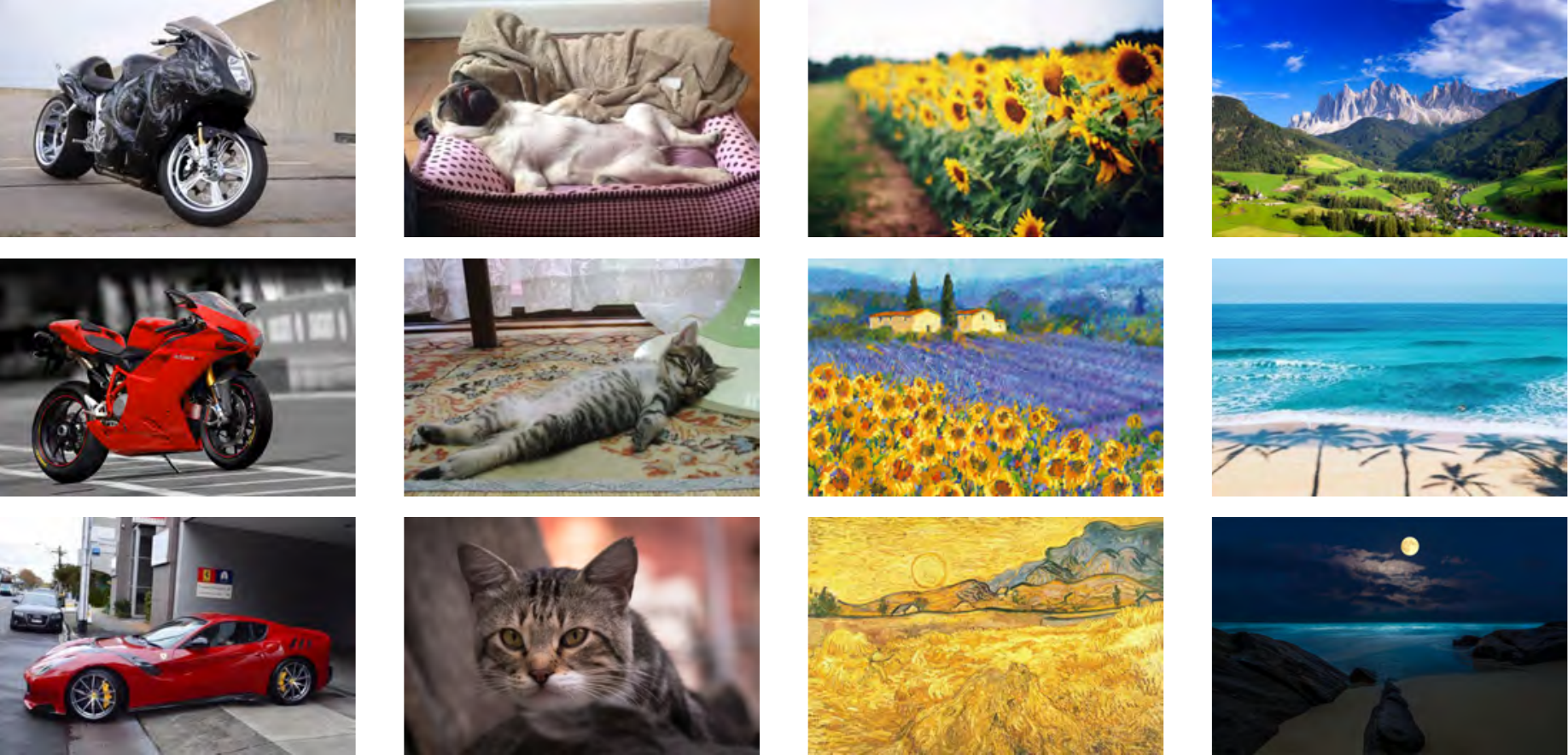}
  \caption{Often the semantic of images is ambiguous. It is unclear whether images in the second row are more similar to the top or bottom row.}
  \label{fig:teaser}
\end{figure}

Understanding semantic similarity among images is the core of a wide range of computer vision analysis and image retrieval applications. However, it is a particularly challenging task as it reflects how humans perceive images, a task that cannot be inferred by low-level analysis. Supervised learning is a common means of studying such semantic problem, for which, the ground truth of how humans perceive similarity among images is critical.

However, the semantic context of images is often ambiguous as images can be perceived with emphasis on different aspects (see Figure~\ref{fig:teaser}). One example out of many is the separation of content and style. One can claim that two images are similar due to their content and another may find two images of similar content different due to their common style.
Similarities between the images may be measured in multiple aspects, which can be contradictory to each other. The key to resolve such contradictions is to disentangle similarities based on their latent aspects.

Humans cannot state a meaningful quantitative measure of semantic similarity. 
\begin{figure}[htbp]
	\centering
	\includegraphics[width=1.0\linewidth]{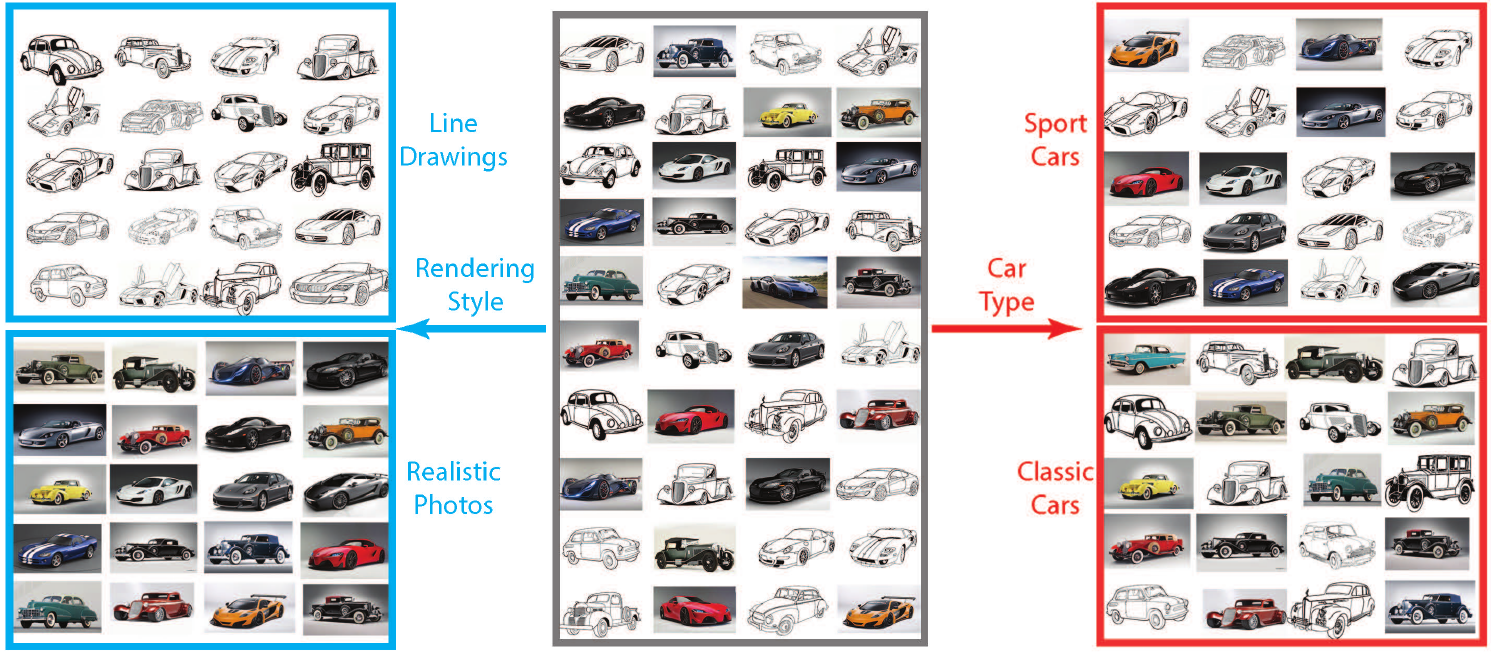}
	\caption{Different clusterings reflect similarity in different aspects, which may be contradictory to each other.}
	\label{fig:twoembeds}
\end{figure}
Therefore, annotations about semantic similarity collected by crowd queries are qualitative in nature.
In addition, they only contain a partial view of a whole dataset. 
To consolidate the partial and possibly conflicting views, the collected qualitative data is often embedded into a common Euclidean space that hosts all the images, such that the quantitative metric distances among images reflect the aggregate of the qualitative measures as much as possible~\cite{Kleiman:2016,Tamuz:2011}. However, since semantic similarity may reflect various aspects, one embedding space cannot represent well multiple distances among images.

In this paper, we present a method for multi-aspect embedding, which disentangles similarity annotations based on their aspects and embeds them into multiple corresponding embedding spaces. The distances in each embedding space well represent object similarity under the corresponding aspect. The task is challenging since it encapsulates a two-fold problem. First, the similarity among images has no clear quantitative measure and thus, it must be deduced from qualitative measures. Second, the aspect that each crowd member relies on is unknown.

Our general approach to the problem is similar to the one presented by Amid and Ukkonen~\cite{Amid:2015}. Qualitative queries are crowdsourced, and the answers to these qualitative queries are embedded into multiple embedding spaces. The multi-embedding is computed by optimizing an objective function that evaluates the embedded distances with respect to the qualitative queries. Each embedding represents the semantic similarities among images using one specific aspect.

A critical issue in the optimization is to infer which aspect is used in answering a particular query. Thus, each query is associated with an additional variable on top of the unknown coordinates of the embedded elements.
The key idea of our approach is to collect and embed qualitative measures in \emph{bundles}. The bundled measures necessarily share the same aspect, which significantly reduces the number of unknown variables. This directly leads to a significant increase in the accuracy of the embeddings.

More specifically, rather than collecting pairs or triples of relations one by one, the task we use is designed as classifying a collection of images into clusters (see Figure~\ref{fig:twoembeds}). This necessarily leads the user to use a single aspect in providing a series of qualitative measures on the collection of images. Each clustering annotation is then converted into a bundle of $T(i, j, \theta)$-like tuples, where $\theta$ indicates whether image $O_i$ is similar to image $O_j$ ($\theta=1$) or not ($\theta=0$), and fed into our bundle embedding optimization. As we shall show, the classification based task is much more cost-effective, and the optimization with tuple bundles requires less variables, leading to higher quality embeddings.

We evaluate our approach on various synthetic and crowdsourced datasets, and show that it significantly outperforms state-of-the-art multi-embedding approaches.

\section{Related Work}
Together with the availability of massive data, crowdsourcing allows supervised learning to be performed in a truly large scale~\cite{Russakovsky:2015}, providing an efficient way to measure human perception in various contexts, such as product design~\cite{Bell:2015}, illustration style~\cite{Garces:2014}, font similarity~\cite{O'Donovan:2014} and entity matching~\cite{Wang:2012}.
For a recent comprehensive study of crowdsourcing, please refer to~\cite{Chittilappilly:2016}.

\vspace{.5em}
\noindent\textbf{Single Embedding Metric Learning.} The problem of consolidating numerous instances of information, which is often quantitative, into a single consistent space is referred to as \emph{metric learning}, and is widely studied~\cite{Globerson:2005,Wang:2011,Xie:2013,Xing:2002}.
Quantifying human similarity perception is challenging since it is often qualitative and relative.
A number of metric learning methods focus on recovering a single embedding space from such relative similarity measures, in the form of paired comparisons~\cite{Agarwal:2007} or relative triplets~\cite{Tamuz:2011}. Kleiman~\etal~\cite{Kleiman:2016} proposed clustering queries which provide more information compared to pairs or triplets of images.

Such methods assume that similarity between two objects can be depicted by a single scalar value, thus a single embedding space can capture similarity among a set of objects. Similarity measures, which might be from different aspects, are ``fused'' into one embedding space. Instead, we model similarity between two objects as a multi-dimension vector, \ie, two objects may have different degree of similarity under different aspects. We propose to ``disentangle'' similarities and embed them in multiple embeddings by their aspects, which can be separately explored.

\vspace{.5em}
\noindent\textbf{Multi-aspect Embedding.} Learning multiple embeddings in general cases has not been explored much, even though it is often essential for various human-computational applications. Recent research in natural language processing has proposed a number of models in which
words are associated with several corresponding embeddings based on human word similarity judgments~\cite{Li:2015,Liu:2015,Wu:2015}. However, these models use additional information such as local co-occurrence and sentence context which are not available in the general case.

A deep model to learn multi-aspect similarity is proposed by Veit~\etal~\cite{Veit2017}. In this work, multi-aspect embeddings are learned directly from image features through a supervised way: the network is provided with a given set of triplets and their corresponding aspect.
Different to this work, none of the data is labeled with accurate aspect.
The gist of our work is to estimate which similarity triplet is associated with each aspect, and at the same time generate multi-aspect embeddings that fit each of these unknown aspects.

Perhaps most similar to our work is the recent work by Amid and Ukkonen~\cite{Amid:2015}. Their multi-view triplet embedding algorithm (MVTE) aims to reveal multiple embeddings by maximizing the sum of log-probabilities of triplet-embedding mapping over all triplets. The triplets are collected individually and it is unknown whether a pair of triplets are induced by the same aspect or different aspects.
The triplet-embedding mapping is defined as an heuristic indicator function of the embedding based on distribution assumptions of the underlying embedding spaces. The method alternates between optimizing the embeddings with fixed indicators and deriving the indicators from embeddings.

Unlike MVTE, we collect image similarity annotations in the form of bundles. Each bundle contains information regarding many images, that share a single unknown aspect. We do not make any distribution assumption of the underlying embedding spaces. Instead, we introduce a set of \emph{aspect inference variables} that represent the mapping probabilities.
These aspect inference variables are optimized simultaneously with the embedding variables, to model complex bundle-embedding relationships.
In Section~\ref{sec:results}, we present a qualitative and quantitative comparison between MVTE and our multi-aspect embedding approach.
\section{Bundle-based Queries}
\label{sec:method}
In this section, we describe our design for collecting crowdsource annotations in bundles which share a common aspect. In the following section, we describe how these bundled qualitative queries are embedded in corresponding multiple quantitative metric spaces.

The design of annotation task is the key for gathering information from a crowd.
For a set of images $\mathcal{S}=\{O_i\}$, a typical task for acquiring similarity annotations is in the following form: is image $O_i$ more similar to $O_j$ or $O_k$? The answer to the question is an ordered triplet: $T(i,j,k)$, in which $O_i$ is more similar to $O_j$ than $O_k$.
Such triplet queries do not scale well to handle large datasets because of their cost-inefficiency.
Moreover, they do not provide context to the crowd worker, which can make the query more difficult to answer, and reduces the consistency of answers among multiple queries. It is possible that each query is answered with a different aspect in mind.

Each triplet query has to be associated with the correct embedding space to produce consistent embeddings. Thus, it is beneficial to use queries which gather similarity information for many triplets using the same aspect. To this end, inspired by~\cite{Kleiman:2016}, we ask workers to perform a clustering query. More specifically, a set of images is presented to each worker, who is requested to classify the images into multiple groups. Note that this classification task is cost-effective since it is only slightly harder than $T(i,j,k)$-like queries, but it can yield a large amount of $T(i,j,k)$-like annotations from these clusters.

More importantly, such a classification naturally leads the worker to use a single similarity aspect while performing the task. Thus, all the derived $T(i,j,k)$ triplets can be assigned to the same embedding space, i.e., they are bundled. The bundle queries greatly reduce the amount of affiliations to be inferred, as instead of inferring the affiliation of each triplet, only a single affiliation for each bundle is required.

Formally, in a query, we ask a crowd worker to classify $\mathbb{N}$ images into at most $\mathbb{B}$ bins/clusters $\{\mathcal{S}_{c}\}$. The aforementioned $T(i,j,k)$-like triplets can be derived from the clusters as $\{T(i, j, k)\}$, where,  $O_i, O_j \in \mathcal{S}_x, O_k \in \mathcal{S}_y$ and $x \ne y$, i.e., two images from the same cluster are considered to be more similar than the third one from another cluster.

In practice, we chose to use a simpler representation of qualitative similarities --- $T(i, j, \theta)$-like tuples. These tuples can be derived from the clusters $\{\mathcal{S}_{c}\}$ by producing a tuple $\{T(i, j, 1)\}$ where $O_i, O_j \in \mathcal{S}_x$ and a tuple $\{T(i, j, 0)\}$ where $O_i \in \mathcal{S}_x, O_j \in \mathcal{S}_y$.
In other words, two images are considered to be similar/dissimilar if they are from same/different clusters.
We denote tuples derived from query $q \in \mathcal{Q}$ as $\mathcal{T}^q = \{T(i, j, \theta)\}$. In the next section, we present a bundle optimization algorithm that takes bundled tuples $T(i, j, \theta)$ as input.

\section{Multi-aspect Embedding in Bundles}

As discussed above, a multitude of aspects cannot be captured in a consistent way within a single embedding space.
Thus, we compute multiple embedding spaces $\mathcal{E} = \{E_s\}$ dedicated to different similarity aspects.

To associate bundled $T(i, j, \theta)$-tuples with appropriate embedding spaces, there are two sets of variables to solve. One set contains the aspect inference variables ${\alpha}^q_s$, which indicates the likelihood that query $q$ is based on the $s$-th similarity aspect. The other set contains the coordinates of the images in each embedding.

Let us denote the coordinates of image $O_{\ast}$ in embedding $E_s$ as $O_{\ast,s}$.
We use constrastive loss~\cite{chopra2005learning} to model how well tuple $T(i, j, \theta)$ fits in the $s$-th embedding $E_s$:
\begin{equation}
\begin{split}
L(T(i, j, \theta),E_{s})= \theta \times {d(O_{i,s}, O_{j,s})}^2  \\
+ (1-\theta) \times {\max(0, m - d(O_{i,s}, O_{j,s}))}^2,
\end{split}
\end{equation}
where $d(O_{i,s}, O_{j,s}) = {\| O_{i,s} - O_{j,s} \|}_2$ and $m$ is a margin for embedding dissimilar images apart from each other.
The loss of associating bundled tuples $\mathcal{T}^q$ with embedding $E_s$ is then:
\begin{equation}
L(\mathcal{T}^q,E_{s})=\sum_{T(i, j, \theta) \in \mathcal{T}^q}{L(T(i, j, \theta),E_{s})}.
\end{equation}

Intuitively, $L(\mathcal{T}^q,E_{s})$ is small when the tuples $\mathcal{T}^q$ from query $q$ are associated with the embedding space that corresponds to the similarity aspect used by query $q$. However, it is unknown which embedding space is the best fit. We introduce aspect inference variables ${\alpha}^q_s$ to address this problem. Formally, the aggregate loss of bundled tuples $\mathcal{T}^q$ with respect to multiple embeddings $\{E_s\}$ is:
\begin{equation}
L(\mathcal{T}^q)=\sum_{E_s \in \mathcal{E}}{\alpha^q_s \times L(\mathcal{T}^q,E_{s})}, \quad \sum^{|\mathcal{E}|}_{s=1}{\alpha^q_s} = 1, \alpha^q_s > 0.
\end{equation}
An inference variable ${\alpha}^q_s$ can be interpreted as the probability that query $q$ is based on the $s$-th similarity aspect. As the optimization progresses, ${\alpha}^q_s$ gradually converge to associate query $q$ with a specific embedding.

Finally, we sum the loss for all queries, and the optimization can be written as:
\begin{equation}
\argmin_{\alpha^q_s, \;\; O_{\ast,s}} \sum_{q \in \mathcal{Q}}\sum_{E_s \in \mathcal{E}} \alpha^q_s \times \sum_{T(i, j, \theta) \in \mathcal{T}^q} L(T(i, j, \theta),E_{s}),
\label{eq:bundle_opt}
\end{equation}
where $\sum_{s}{\alpha^q_s} = 1$, and $\alpha^q_s > 0$. Note that there is one aspect inference variable $\alpha^q_s$ per query per embedding, i.e., their total number is $ |\mathcal{Q}| \times |\mathcal{E}|$.  The loss function is differentiable with respect to variables $\alpha^q_s$ and $O_{\ast,s}$, thus it can be optimized with gradient descent based optimizers. We solve these two sets of variables simultaneously.

\vspace{.5em}
\noindent\textbf{Initialization.}
If the embedding spaces are initialized with the same coordinates, or symmetrically with respect to the queries, the gradients are exactly the same. Thus, the gradient descent optimization updates them in the same way, which leads to identical embeddings. To avoid this, we use random initialization for the embedding coordinates.
It can be assumed that the initial random embeddings are not equivalent or symmetric to one another with respect to the queries. However, in the beginning, such non-equivalence/asymmetry is probably weak, i.e., there is no strong tendency for a query to belong to a specific embedding. The non-equivalence/asymmetry is gradually reinforced by our algorithm, and the embeddings evolve into quite different spaces corresponding to multi-aspects.

We initialize $\alpha^q_s$ to $\frac{1}{|\mathcal{E}|}$, indicating that the queries have the same probability to be based on any of the unknown aspects. For specific applications, where relevant prior information can be leveraged, they can also be initialized with bias for different aspects.
\begin{figure*}[htbp]
	\centering
	\includegraphics[width=0.89\linewidth]{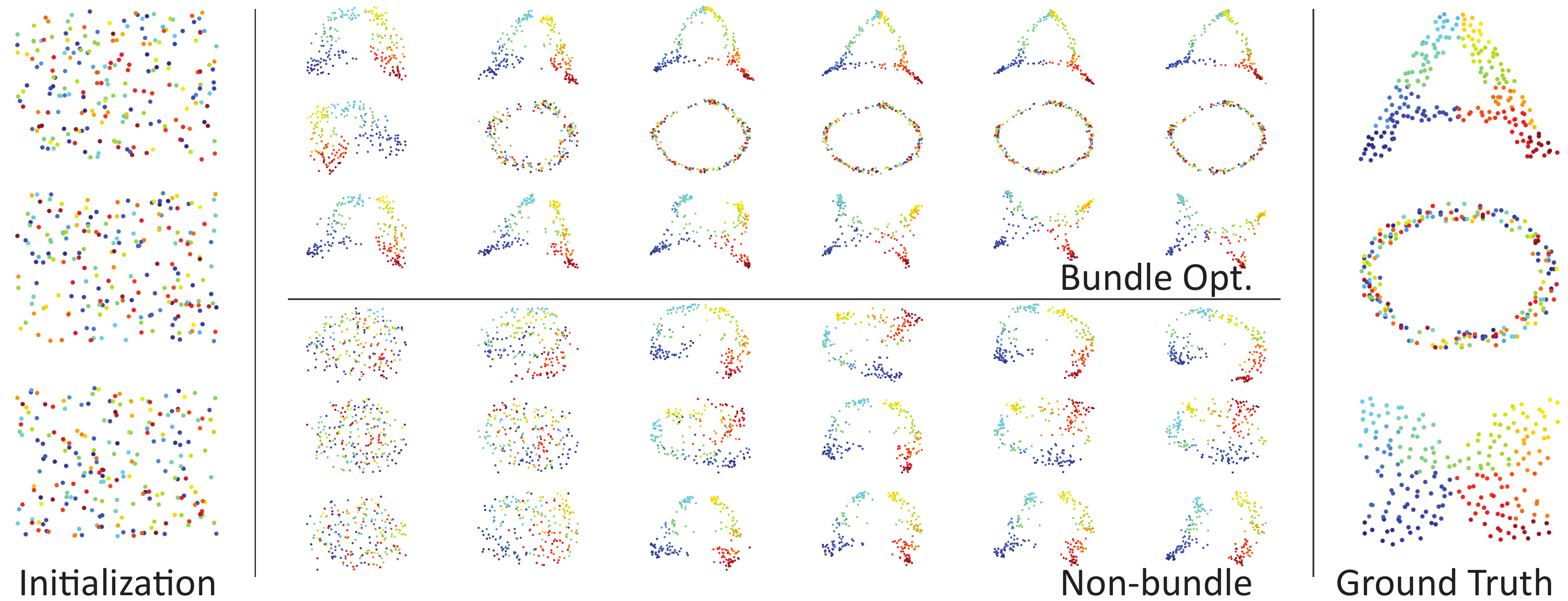}
	\caption{A visual comparison of embedding results w/wo bundle optimization at iteration 5, 10, 20, 30, 60 and 100.}
	\label{fig:bundleiters}
\end{figure*}

\section{Algorithm Analysis}
\label{sec:analysis}
Two distinctive features of our algorithm are the bundle optimization and the use of aspect inference variables. We study their effectiveness by comparison to alternatives approaches.

\vspace{.5em}
\noindent\textbf{Analysis Settings.} The goal of our algorithm is to estimate semantic similarity among images based on human perception. This similarity measure can be used as ground truth for supervised machine learning algorithms.
However, the evaluation of our algorithm cannot be based on crowdsourcing data which is inexact in nature.
Instead, we use synthetic data as ground truth to analyze and evaluate our algorithm.
\begin{figure}[t]
	\centering
	\includegraphics[width=1.0\linewidth]{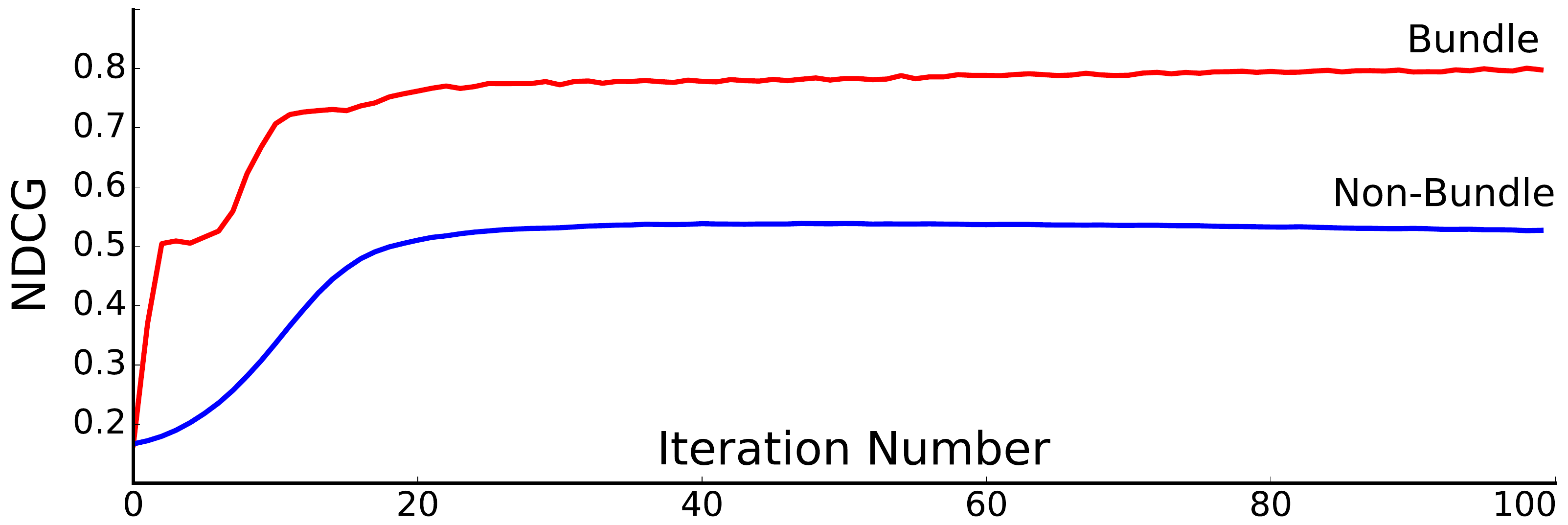}
	\caption{A comparison of Fig.~\ref{fig:bundleiters} results in NDCG metric.}
	\label{fig:bundle_ndcg}
\end{figure}

We introduce a synthetic ``AOB'' dataset, which contains 214 points ($|O_{\ast}| = 214$),
distributed to form ``A'', ``O'', and ``B''utterfly shapes in the ground truth embeddings $\mathcal{E}_{gt} = \{E_{gtA}, E_{gtO}, E_{gtB}\}$ (see Figure~\ref{fig:bundleiters}).
The points are indexed sequentially according to point coordinates in $E_{gtA}$ and $E_{gtB}$, so the embeddings are different but not completely independent. In $E_{gtO}$, the points are indexed randomly, so that the embedding is completely independent of the other two. This way, the dataset simulates both dependent and independent aspects. We color the points by smoothly mapping their indices into continuous colors, \ie, points with neighboring indices have similar colors.
\begin{figure}[t]
	\centering
	\includegraphics[width=1.0\linewidth]{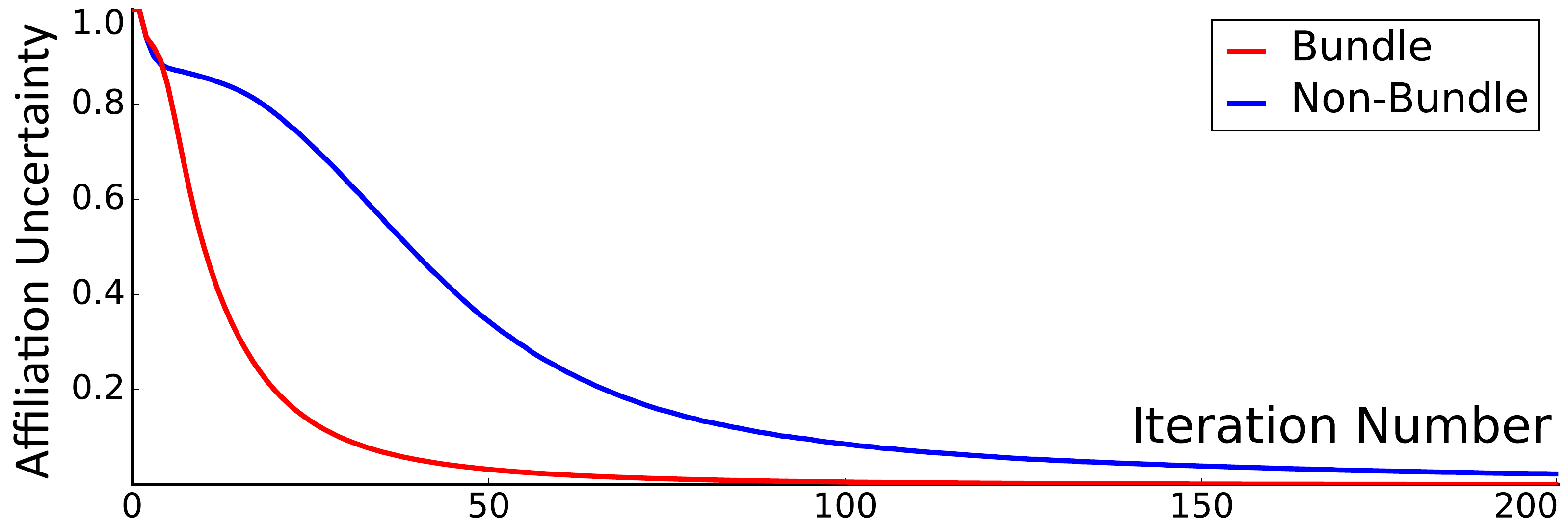}
	\caption{Affiliation uncertainty curve during recovery of ``AO'' embeddings, with and without bundle optimization.}
	\label{fig:bundle_uncertainty}

\end{figure}

We generate $|\mathcal{Q}| (= 600)$ random queries from each ground truth embedding and attempt to recover them by simulated query answers.
Each query contains $\mathbb{N}(=20)$ randomly sampled objects. Note that the random sampling strategy does not use any prior knowledge of the embeddings, to simulate actual crowdsourcing scenario where the ground truth is unknown. The answers are generated using K-means clustering of the samples, with $\mathbb{B}(=5)$ seeds. The clustering is based on the position of objects in one of the embeddings (selected in random), to simulate users query answers which are based on a single unknown aspect. $114,000$ tuples are inferred from the clustering query answers of each embedding.

We evaluate the quality of recovered embeddings $\mathcal{E}$ based on the Normalized Discounted Cumulative Gain (NDCG) metric~\cite{Jarvelin:2000}. NDCG is widely used in evaluating retrieval relevance, and suitable for evaluating the recovery quality of the similarity based embedding spaces. More specifically, we first compute K-nearest ($K=0.1 \times |O_{\ast}| $) neighbors for each point in each recovered embedding in $\mathcal{E}$ and its corresponding embedding\footnote{In case of poor recovery, the correspondence between $\mathcal{E}$ and $\mathcal{E}_{gt}$ is not clear (see Figure~\ref{fig:bundleiters} middle lower). In this case, we compute NDCG for all possible mappings between them, and pick the one with the highest NDCG as the most likely mapping.} in $\mathcal{E}_{gt}$, then with the corresponding ranked lists, NDCG are computed and averaged over all points.

\subsection{Bundle Optimization}
A common approach for computing multi-embeddings is to collect and optimize over individual tuples. In this approach, each tuple may be based on a different aspect, thus the optimization has to infer aspects for each tuple individually. Formally, the non-bundled optimization problem can be written as:
\begin{equation}
\argmin_{\alpha_s^{(i,j,\theta)}, \;\; O_{\ast,s}} \sum_{E_s \in \mathcal{E}} \sum_{T(i, j, \theta) \in \mathcal{T}} \alpha_s^{(i,j,\theta)} \times L(T(i, j, \theta),E_{s}),
\label{eq:non_bundle_opt}
\end{equation}
\begin{figure}[t]
	\centering
	\includegraphics[width=.73\linewidth]{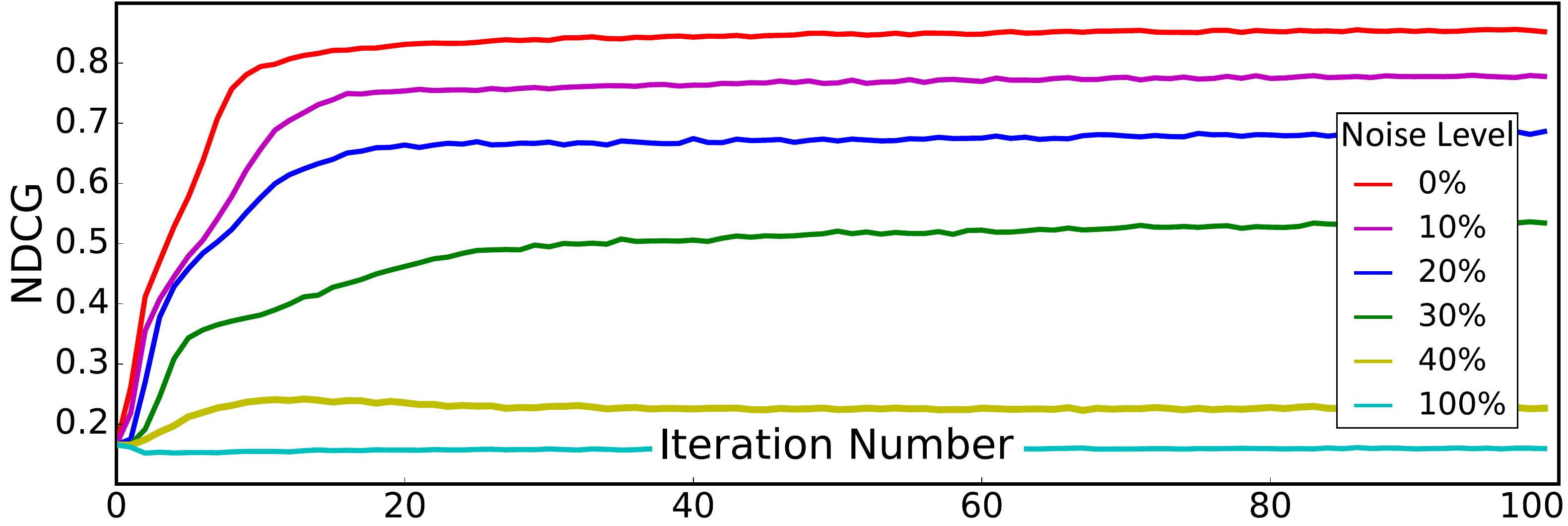}
	\caption{NDCG curve during the recovery of ``AO'' embeddings, with various levels of noise.}
	\label{fig:noise}

\end{figure}
where $\alpha_s^{(i,j,\theta)}$ is the aspect inference variable indicating the probability of associating tuple $T(i, j, \theta) \in \mathcal{T}$ with embedding $E_s$, and $\sum_s{\alpha^{(i,j,\theta)}_s} = 1$, and $\alpha^{(i,j,\theta)}_s > 0$. The tuples can either be collected using single-tuple queries or inferred from a clustering query. Clearly, this formulation leads to many more variables to optimize than the bundled optimization, since the number of aspect inference variables is proportional to the number of tuples.

We apply bundled and non-bundled optimizations on AOB dataset, and show a visual comparison in Figure~\ref{fig:bundleiters}. Note that in both optimizations the embeddings are computed from the same random initialization. As can be seen in the figure, bundled optimization leads to a significantly better recovery of the ground truth embeddings than the non-bundled version. The bundled optimization (top row) produces distinct embeddings that resemble the ground truth, while the non-bundled optimization (bottom row) produces noisy embeddings that are quite similar to each other.
The recovery quality is also evident in the color coding of the results. A high quality recovery should present color coding which is similar to the ground truth. While this is true for the bundled optimization results, in the non-bundled optimization results the color coding of all embeddings is similar only to the ``A'' and butterfly shapes and not to the ``O'' shape. This suggests that the aspects are not separated correctly, as all embeddings are influenced by tuples that represent similar color coding.

We also quantitatively measure the quality of the multi-aspect embedding recovery. At each iteration, we compute the average NDCG of the three embeding results, with bundled and non-bundled optimization. As can be seen in Figure~\ref{fig:bundle_ndcg}, the bundled optimization converges much faster to more accurate embeddings.

\subsection{Aspect Inference Variables}
As discussed above, the optimization starts with random initialization of multiple embedding spaces, and progressively evolves to differentiate between distinct aspects.
We examine this phenomenon in the task of recovering $E_{gtA}$ and $E_{gtO}$ from simulated queries and answers, with an affiliation uncertainty metric:
\begin{equation}
\mathscr{U} = \frac{1}{|\mathcal{Q}|}*\sum_{q \in \mathcal{Q}}{\frac{\min(\alpha_1^q, \alpha_2^q)}{\max(\alpha_1^q, \alpha_2^q)}},
\label{eq:uncertainty}
\end{equation}
where $\alpha_1^q$ and $\alpha_2^q$ are the aspect inference variables associated with $E_A$ and $E_O$ in query $q$. As shown in Figure~\ref{fig:bundle_uncertainty}, in the beginning, $\mathscr{U}$ is high, as the two initial embeddings are still in chaos state and it is not significant whether a query is associated with $E_A$ or $E_O$. However, since $E_A$ and $E_O$ are not likely to be symmetric with respect to the queries, the asymmetry is gradually reinforced while one of the embeddings is evolving towards $E_A$ and the other one towards $E_O$. This can be observed from the reduction of affiliation uncertainty as the optimization progresses.
\begin{figure}[t]
	\centering
	\includegraphics[width=.9\linewidth]{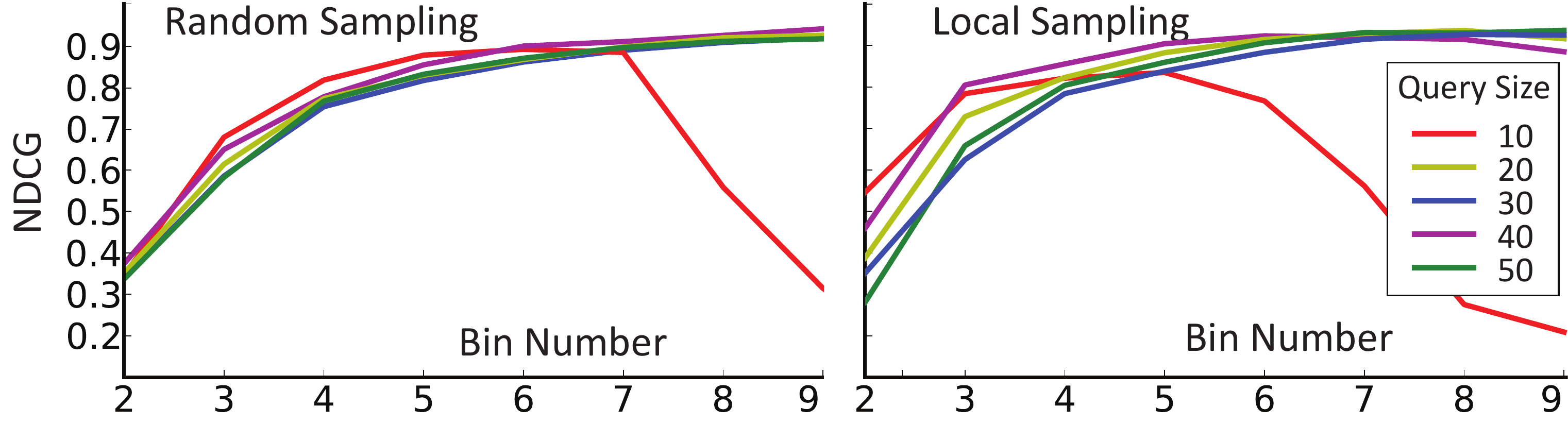}
	\caption{NDCG curve during the recovery of ``AO'' embeddings, using different query sampling strategy, query size and bin number.}
	\label{fig:parameters}
\end{figure}

Similarly, we can define the affiliation uncertainty metric for non-bundled optimization as:
\begin{equation}
\mathscr{U}_n = \frac{1}{|\mathcal{T}|}*\sum_{T(i, j, \theta) \in \mathcal{T}}{\frac{\min(\alpha_1^{(i, j, \theta)}, \alpha_2^{(i, j, \theta)})}{\max(\alpha_1^{(i, j, \theta)}, \alpha_2^{(i, j, \theta)})}},
\label{eq:uncertainty_non}
\end{equation}
where $\alpha_1^{(i, j, \theta)}$ and $\alpha_2^{(i, j, \theta)}$ are the aspect inference variables associated with $E_A$ and $E_O$ for tuple $T(i, j, \theta)$. $\mathscr{U}_n$ is also plotted in Figure~\ref{fig:bundle_uncertainty}. $\mathscr{U}_n$ also reduces as the optimization progresses, which shows the aspect inference variables are somewhat effective without the bundled optimization. Still, the bundled optimization reduces affiliation uncertainty more effectively than the non-bundled version.

\subsection{Stress Test under Noise}
To test the robustness of our method to noise and erroneous measures we performed the following test. We collected the $228,000$ tuples from the clustering query answers of ``A'' and ``O'' embeddings, which form ground truth tuples. Then we add noise to the ground truth tuples by randomly selecting tuples and reversing their similar/dissimilar label. The portion of reversed tuples defines the noise level.

Figure~\ref{fig:noise} summarizes the results, where we can observe that our method tolerates small amount of noise, and the performance degrades smoothly. This indicates that the bundled optimization method is robust to reasonable degree of noisy measures.
\begin{figure*}[htbp]
	\centering
	\includegraphics[width=.95\linewidth]{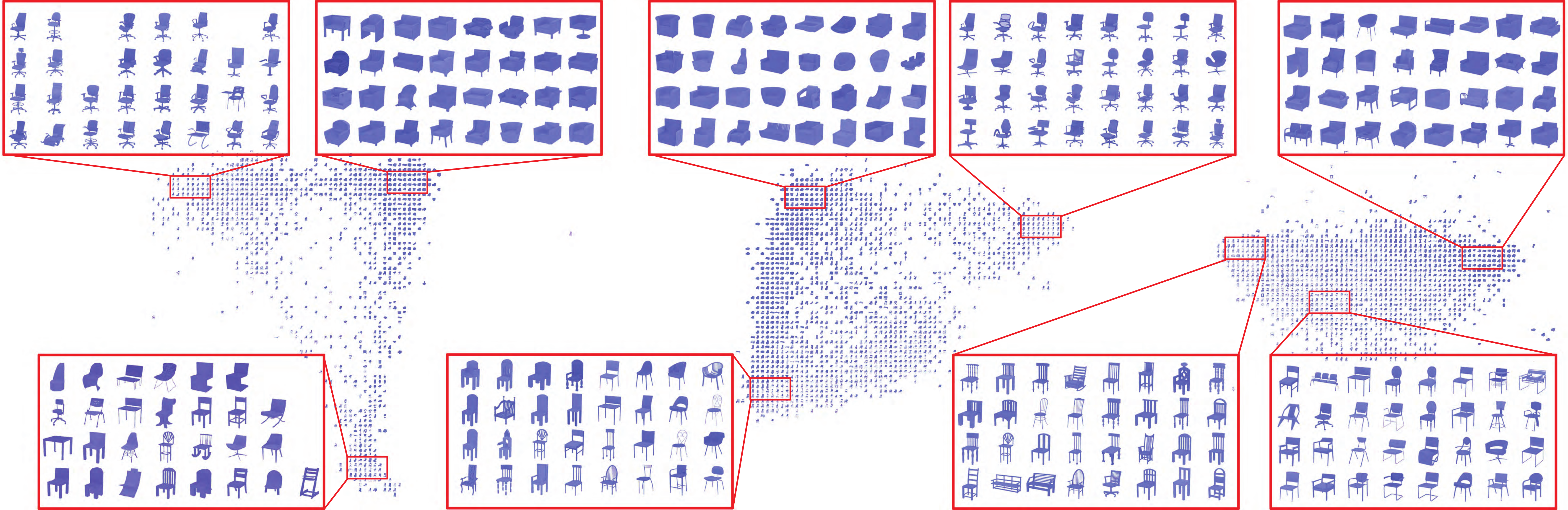}
	\caption{Recovering multiple aspects using our approach on the chair dataset with three aspects. The results show that the embeddings reflect specific aspects, from left to right: arms, legs and back of and chairs.}
	\label{fig:chairembeds}
\end{figure*}
\begin{figure*}[htbp]
	\centering
	\includegraphics[width=.95\linewidth]{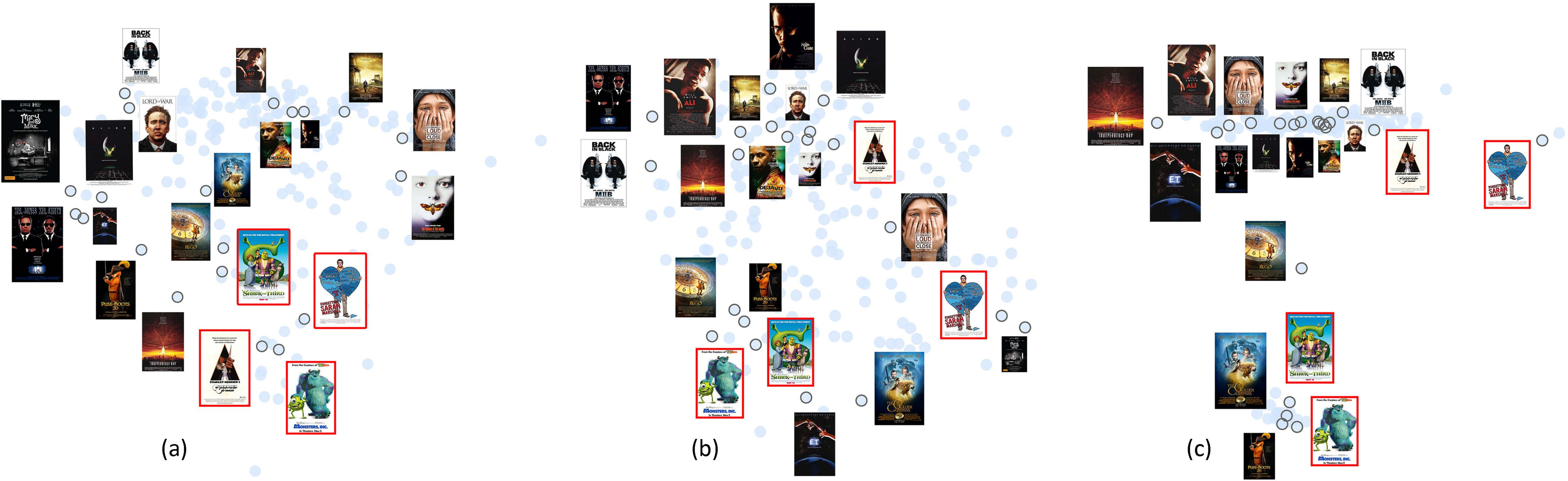}
	\caption{Recovering multiple aspects using our approach on the poster dataset with unknown aspects. The results show that the embeddings are ordered under different attributes: color and appearance (a), genre (b), animation / live action (c).}
	\label{fig:posterembed}
\end{figure*}
\subsection{Query Sampling, Query Size, and Bin Number}
\label{sec:sample}
There are three factors controlling how we present each query to users: query sampling strategy, query size $\mathbb{N}$ and bin number $\mathbb{B}$. Query sampling controls how the images clustered are sampled from the entire set in each aspect, either randomly or locally (as proposed in~\cite{Kleiman:2016}). In the local sampling strategy, the recovery is divided into several phases. Each phase first updates the embedings, and then sample points to form queries from neighboring regions, rather than randomly, in the following phase. $\mathbb{N}$ is the number of images to be clustered in each query. $\mathbb{B}$ is the number of bins --- the maximum number of clusters allowed for each query. We evaluate the performance of our bundle optimization with synthetic data generated with different sampling strategy, $\mathbb{N}$ and $\mathbb{B}$.

We summarize the results in Figure~\ref{fig:parameters}. Note that, to make fair comparisons, we make sure $|\mathcal{Q}| \times \mathbb{N} = 3000$, i.e., more queries were used in tests with smaller query size. We can see that larger $\mathbb{N}$ and $\mathbb{B}$ give better recovery results, which is not surprising, since more information can be collected from such queries. However, the NDCG gradually goes down with larger $\mathbb{B}$ when $\mathbb{N} = 10$, since our approach degrades to the non-bundle case.
On the other hand, smaller $\mathbb{N}$ and $\mathbb{B}$ are more friendly to crowd workers. In particular, a small value of $\mathbb{B}$ encourages workers to make decisions based on a single aspect in each query. We also show that the local sampling strategy proposed in~\cite{Kleiman:2016}, when extended to multiple-aspect setting, is more effective with small query sizes and bin numbers.
There exists a sweet point which balances these factors and maximize the cost-effectiveness.

\section{Experimental Results}
\label{sec:results}
In this section, we provide a dataset with clustering labels of three known aspects and a dataset with clustering labels of unknown aspects, both of which are collected from crowd workers via Amazon Mechanical Turk (AMT). We also compare our results with relevant previous work for learning single or multiple aspects on benchmark datasets.

Our optimization is implemented with Tensorflow~\cite{tensorflow2015-whitepaper}. In Equation~\ref{eq:bundle_opt} and~\ref{eq:non_bundle_opt}, instead of directly optimizing $\alpha_s$ to satisfy $\sum_{s}{\alpha_s} = 1$ and $\alpha^q_s > 0$, we let $\alpha_s = softmax(\beta_s)$, and optimize $\beta_s$ without constraints. We use Adam method~\cite{kingma2014adam} with learning rate $0.01$ for the optimizations. All code and data will be opensourced.\\

\subsection{Results with Bundles Collected from AMT}
In this subsection, we use AMT crowdsourced data (either with clustering labels of known or unknown aspects) to learn multi-aspect embeddings. Local sampling strategy (see Section~\ref{sec:sample}) proposed in~\cite{Kleiman:2016} was adopted to produce clustering queries.
\label{sec:amt}

\begin{figure*}[t]
	\centering
    \includegraphics[width=.95\linewidth]{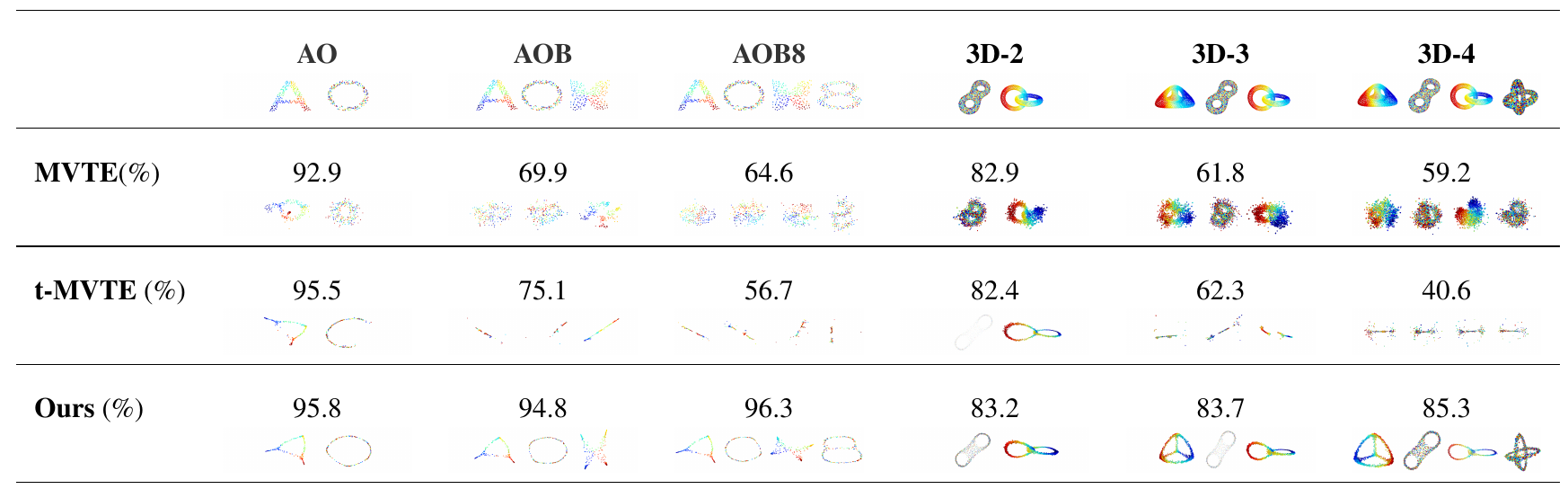}
	\caption{NDCG performance and embedding visualization compared with (t-)MVTE algorithms on various datasets.}
	\label{fig:ndcgcmp}
\end{figure*}
\begin{figure*}[t]
	\centering
    \includegraphics[width=.95\linewidth]{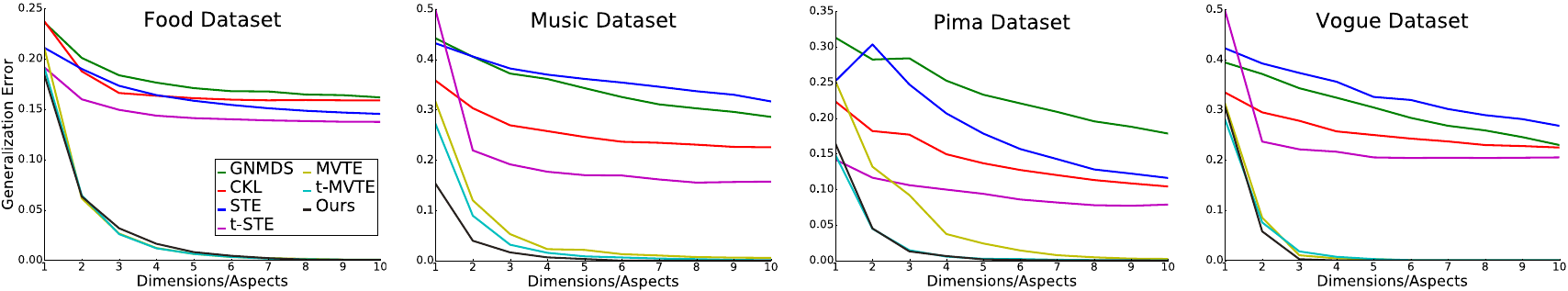}
	\caption{Comparison with state-of-the-art single and multiple embedding methods on Food, Music, Pima and Vogue datasets.}
	\label{fig:compchart}
\end{figure*}
\vspace{.5em}
\noindent\textbf{Recovering predefined aspects.}
To properly evaluate the performance of our method, we require a dataset that is intuitive enough for crowd queries, but also relates to known predefined aspects.
To this end, we use a chair dataset which contains $6,777$ images rendered from the chair category of ShapeNet~\cite{shapenet2015} and collect user data which related to multiple aspects in this dataset.
Crowd workers were required to cluster queries considering one of the following predefined aspects: arms, legs and back of the chairs (please refer to our appendix for AMT experiment details).
We then recover appropriate embeddings of these aspects \emph{without} the prior knowledge of the aspect the users were asked to consider.
Figure~\ref{fig:chairembeds} shows the final recovered embeddings for three aspects. It can be seen that each embedding reflects a separate aspect. This demonstrates the ability of our method to automatically distinguish and classify the query answers according to their unknown aspects.

\vspace{.5em}
\noindent\textbf{Recovering unknown aspects.}
In this experiment, we use our multi-aspect embedding method to embed movie poster images~\cite{Kleiman:2016} using AMT queries which do not have any predefined aspect. That is, the crowd workers are not guided to use specific aspects, but can choose any aspect they see fit when clustering the images.
We then compute three-aspect embeddings using our method. The results can be seen in Figure~\ref{fig:posterembed}. For the purpose of display clarity, we only present $20$ sampled images out of the total in the dataset. The nature of images in the dataset and the unconstrained settings of the experiment suggest the images may be categorized by the crowd workers using a large number of aspects. Thus, each embedding may reflect a mix of several aspects. Still, we can identify a meaningful distinction between the three embeddings. Embedding (a) reflects the appearance of the posters, in terms of color, composition and the content of the poster. For example, posters with white background (see marked images) appear close to each other in this emebedding, even though the movies belong to various genres.
Embedding (b) reflects external context such as the genre of the movie or the actors that play in it. Note that horror movies appear on the top right, sci-fi movies appear on the top left, and family movies appear on the bottom. In embedding (c), the distinction between animated movies and live action movies takes precedence over other aspects, creating two tight groups of movies, animated and non-animated.

\subsection{Comparisons with State-of-the-Art Methods}
\label{sec:star}
\vspace{1em}
\noindent\textbf{Comparisons with multi-embedding methods.} We conduct an experiment to evaluate the effectiveness of our algorithm and compare it with (t-)MVTE multi-view embedding algorithm~\cite{Amid:2015}. We use six synthetic datasets with multiple aspects, as well as different dimensions, as can be seen in Figure~\ref{fig:ndcgcmp}.
For example, 3D-2 is a set of three-dimensional points composed of two geometric models that correspond to two aspect distributions.
Each 2D data has $214$ instances, while $1600$ instances for each 3D data.
$600$ and $4800$ clustering queries are randomly sampled from each aspect in the synthetic 2D and 3D data, which are used to produce triplets and bundled tuples. Triplets or bundled tuples in each aspect are mixed together as input to (t-)MVTE or our method for optimizing multi-aspect embedding.
Figure~\ref{fig:ndcgcmp} summarizes the NDCG of the recovered multi-embedding compared with the ground truth, and corresponding qualitative visualized embeddings.
Our algorithm outperforms (t-)MVTE, and is more stable when dealing with complex data of higher dimensions.

\vspace{.5em}
\noindent\textbf{Comparisons with single embedding methods.} We also compare our method with several previous single embedding methods, including GNMDS~\cite{Agarwal:2007}, CKL~\cite{Tamuz:2011}, and  (t-)STE~\cite{Maaten:2012} using datasets from previous work: Food~\cite{wilber2014cost}, Music~\cite{ellis2002quest}, Pima~\cite{smith1988using}, and Vogue~\cite{HeikinheimoU13}.
Pima dataset contains 768 instances, each having 8 features indicating people's physical conditions. We adopt the method in ~\cite{Amid:2015} to produce similarity triplets, which generate $100$ triplets for each instance in each aspect. The other three datasets offer individually collected $\{T(i, j, k)\}$ similarity triplets. We further convert each $\{T(i, j, k)\}$ triplet to $\{T(i, j, 1)\}$ and $\{T(i, k, 0)\}$ tuples, which are bundled together.

As in~\cite{Amid:2015}, generalization error is used to evaluate the performances, which describes the dissatisfaction ratio of new recovered triplets in the ground truth. For multi-embedding methods, a triplet is considered to be satisfied if its distance relationship in one of the multi-aspect embedding is consistent with that in ground truth. Since single embedding methods cannot recover multi-aspect embedding, we compare with their ability to recover corresponding high dimensional space. Figure~\ref{fig:compchart} shows that multi-aspect embedding methods outperform single embedding methods, and the tendency is more obvious as dimensions/aspects increase.

Note that, for these datasets, which contain only minimal bundled information --- each triplet is considered as a bundle, our algorithm degrades to non-bundle case.
However, as shown in Figure~\ref{fig:compchart}, even in this case, our method performs comparable with the other methods. Also note that our method does not make any assumption on the underlying data distribution, as that in (t-)MVTE.
In a summary, \emph{our method performs significantly better when bundle information is present, and comparable when only non-bundled triplets are provided}, while making no assumption of the data distribution.

\vspace{.5em}
\noindent\textbf{Limitations.} Different similarity aspects may have different popularity. Similarly to (t)-MVTE, our method does not take this into account. In addition, the number of different embeddings in our method needs to be manually set. While larger number of embeddings can better reflect more aspects, there is a risk that they actually represent noise or outlier measures. An interesting future work is to try and differentiate inliers and outliers measures, and automatically pick a suitable number of embedding spaces.

\section{Conclusions}
\label{sec:conclusions}

We have presented a method for multi-aspect embedding. The method takes qualitative measures and solves an optimization problem that embeds them into multiple spaces such that the quantitative measures in the embeded spaces agree with the qualitative measures. The optimization solves two sets of unknown parameters simultaneously: one is the embedded coordinates of the points and the other is the classification variables of the measure to the unknown aspects. We presented a bundle optimization and showed its power to infer the aspect classification. We showed that it outperforms existing multiple embedding methods.
Our experimental results on crowdsourced data demonstrate the competence of our method to produce multi-embedding from inconsistent and redundant data.
\section*{Acknowledgements}
Our work was supported by National Key Research \& Development Plan of China (No. 2016YFB1001404) and National Natural Science Foundation of China (No. 61602273).


\bibliographystyle{acmtog}
\bibliography{multiembed}


\begin{appendices}
\section{Appendix I: NDCG Calculation}
We evaluate the quality of multi-embedding based on Normalized Discounted Cumulative Gain (NDCG) metric~\cite{Jarvelin:2000,Rvelin2002Cumulated}. NDCG is widely used for evaluating information retrieval, and is suitable for evaluating the recovery quality of an embedding space.
The NDCG for a point $p$ in an embedding space is defined as ${\text{NDCG}}_{p}=\frac{{\text{DCG}}_{p}}{{\text{IDCG}}_{p}}$, and the NDCG of an embedding space is defined as the average NDCG of the points in the space. The term ${\text{DCG}}_{p}$ is the Discounted Cumulative Gain for point $p$, and is defined as:
\begin{equation}
{\text{DCG}}_{p}=\sum_{i=1}^{K}\frac{2^{{rel}_{i}}-1}{log_{2}(i+1))},
\end{equation}
where $K=0.1 \times |O_{\ast}|$ is the number of nearest neighbors in the embedding space to be evaluated, and ${rel}_i$ is the relevance between $p$ and its $i$-th nearest neighbor $p_i$. The relevance is defined as:
\begin{equation}
{rel}_{i}=e^{-\frac{d(p, p_i)}{d(p, p_{\text{\tiny K}})}},
\end{equation}
where $d(p, p_i)$ denotes the distance between $p$ and $p_i$ in the \emph{ground truth} embedding space.
The term $\text{IDCG}_{p}$ is the ideal $\text{DCG}_{p}$, \ie, $\text{DCG}_{p}$ computed in the ground truth embedding space. For a perfectly recovered embedding space, ${DCG}_{p}$ is the same as ${IDCG}_{p}$, producing $\text{NDCG}_{p} = 1$. A higher NDCG value indicates a better approximation of the ground truth embedding.

\section{Appendix II: AMT Data Collections}
\begin{figure}[t]
	\centering
	\includegraphics[width=1\linewidth]{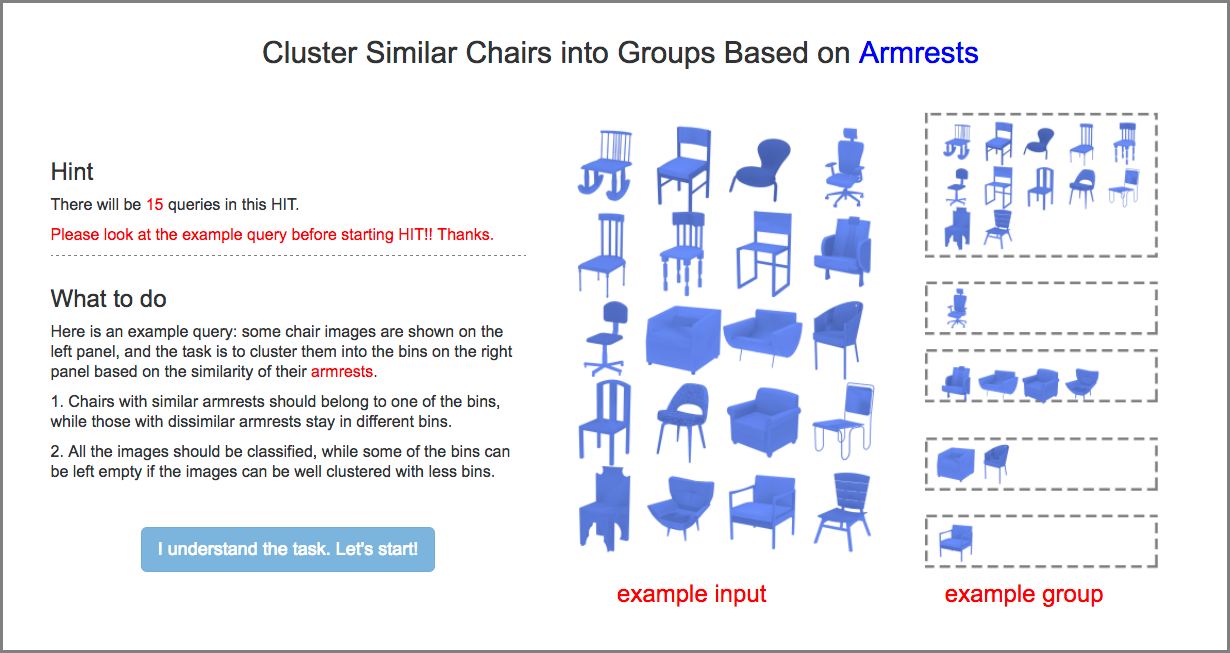}
	\caption{MTurk HIT introduction for predefined-aspect (\eg, arm) experiment on Chair dataset.}
    \label{fig:armhitintro}
\end{figure}

\vspace{.5em}
\noindent\textbf{Predefined-aspects Experiment on Chair Dataset.}
We collected semantic similarity data using clustering queries (produced by $2$ updating phases of local sampling strategy) instead of the more traditional triplet queries, using a drag-and-drop graphical UI with $20$ clustering images shown on the left and $5$ grouping bins on the right.
Workers were required to cluster the $20$ images into the bins, with similar ones in the same bin. An experimental task in AMT for each worker is considered as a human intelligence task (HIT). Each HIT begins with an examplar introduction with guidelines for crowd workers (see Figure~\ref{fig:armhitintro}), followed by $15$ queries.
As a quality control, $3$ queries in each HIT are ground truth sentinels, answers from a worker with lower than $70\%$ sentinel accuracy will be rejected. Additionally, only crowd workers with higher than $80\%$ approval rate can accept our HIT.

\begin{figure}[t]
	\centering
	\includegraphics[width=1\linewidth]{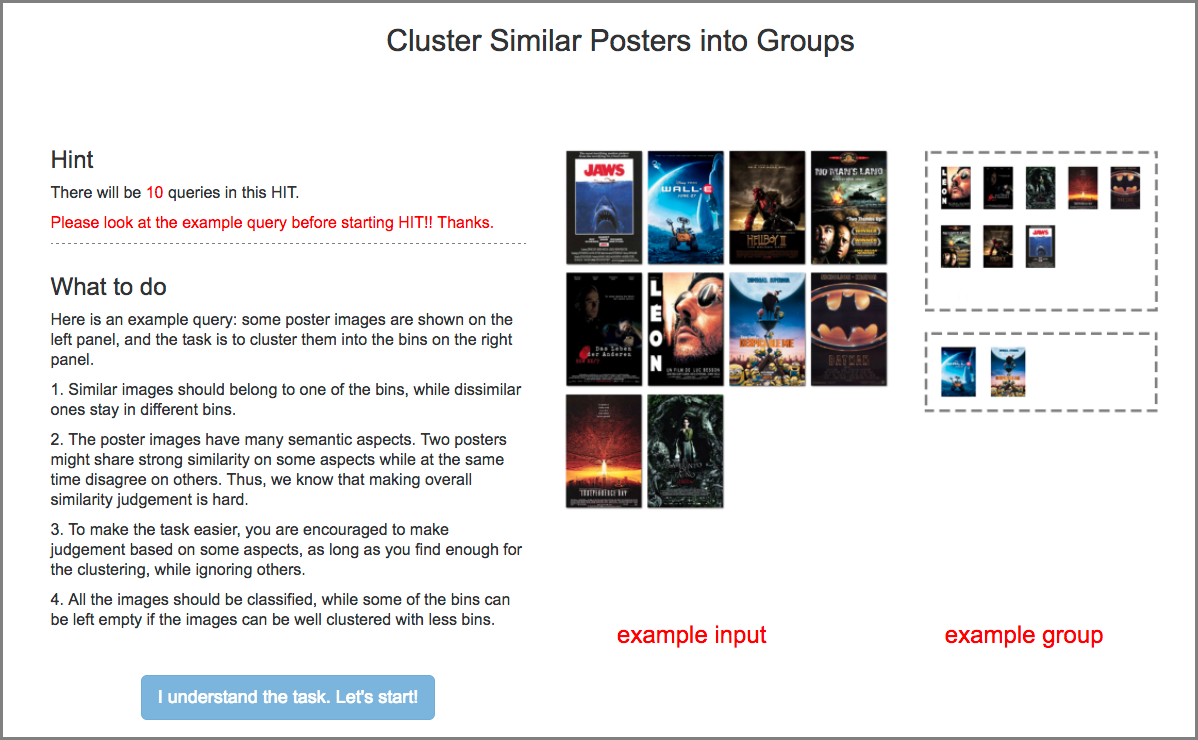}
	\caption{MTurk HIT introduction for unknown-aspect experiment on Poster dataset.}
    \label{fig:posterhitintro}
\end{figure}
We distribute in total 45,000 queries with ground truth sentinels for the predefined-aspects experiment, producing 3,000 HITs. After quality control, we collected 41,287 valid clustering query answers, which are aggregated to 7,953,827 final tuples. Workers are paid on average $\$0.25$ to cluster $15$ queries and spent an average of $6$ minutes per HIT. The total cost was about $\$800$ spent over roughly a month and a half.

\vspace{.5em}
\noindent\textbf{Unknown-aspects Experiment on Poster Dataset.}
Film posters present rich semantic information, which makes it hard for workers to cluster queries without any predefined aspects.
To simplify the task, we collect clustering queries (produced by $1$ updating phase of local sampling strategy) using a drag-and-drop graphical UI with $10$ clustering images shown on the left and $2$ grouping bins on the right.
Workers were asked to cluster the $10$ images into the bins according to their own preferences.
In this experiment, each HIT begins with an examplar introduction with guidelines for crowd workers (see Figure~\ref{fig:posterhitintro}), followed by $10$ queries. Since there is no ground truth answers for this experiment,
we constrain valid crowd workers as those higher than $80\%$ approval rate as a quality control.

In total, we distributed $840$ queries, producing in total $84$ HITs. After excluding incomplete answers, we collected $800$ valid queries which are aggregated to 36,000 final tuples. Workers are paid on average $\$0.25$ to cluster $10$ queries and spent an average of $3$ minutes per response. Please note that the reward was higher in this experiment to accelerate the data collection. The total cost was about $\$24$ spent over one day.

\end{appendices}

\end{document}